\pdfoutput=1
\documentclass[11pt]{article}

\usepackage{acl}

\usepackage{times}
\usepackage{latexsym}

\usepackage[T1]{fontenc}

\usepackage[utf8]{inputenc}

\usepackage{microtype}

\usepackage{mdframed}
\usepackage{xcolor}

\usepackage{booktabs}
\usepackage{graphicx}
\usepackage{multirow}

\usepackage{float}
\usepackage{adjustbox}

\usepackage{enumitem}

\usepackage{tikz,pgfplots,pgfplotstable}
\usetikzlibrary{shapes.misc, positioning, decorations.pathreplacing, calc}
\usepackage{subcaption}

\usepackage{colorblind}
\usepackage{soul}
\DeclareRobustCommand{\hlc}[2]{{\sethlcolor{#1}\hl{#2}}}

\usepackage{tcolorbox}
\definecolor{azure}{rgb}{0.0, 0.5, 1.0}
\tcbuselibrary{skins}
\tcbset{enhanced}

\usepackage{amssymb}

\definecolor{tgb1}{rgb}{0.2666, 0.466, 0.666}
\definecolor{tgb2}{rgb}{0.4, 0.8, 0.9333}
\definecolor{tgb3}{rgb}{0.1333, 0.5333, 0.2}
\definecolor{tgb4}{rgb}{0.8, 0.733, 0.2666}
\definecolor{tgb5}{rgb}{0.9333, 0.4, 0.466}
\definecolor{tgb6}{rgb}{0.666, 0.2, 0.466}

\definecolor{redrag}{rgb}{0.97254902, 0.901960784, 0.929411765}
\definecolor{bleurag}{rgb}{0.941176471, 0.952941176, 0.9529411760}
\definecolor{yellowrag}{rgb}{0.984313725, 0.952941176, 0.905882353}
\definecolor{purplerag}{rgb}{0.956862745, 0.937254902, 0.984313725}

\newcommand{\guillemet}[1]{``#1''}

\usepackage[english]{babel}
\usepackage{hyperref}
\addto\extrasenglish{%
}

\title{Quebec Automobile Insurance Question-Answering With Retrieval-Augmented Generation}

\author{David Beauchemin\thanks{\hspace{4pt}Contributed equally to this work.}~\hspace{1.5pt}\textsuperscript{\textdagger}, Zachary Gagnon\protect\footnotemark[1]~\hspace{1.5pt}\textsuperscript{\textdaggerdbl}, \and Richard Khoury\textsuperscript{\textdagger}\\
Universit\'e Laval, Qu\'ebec, Canada \\
Computer Science Department\textsuperscript{\textdagger}\\
\textsuperscript{\textdagger}\texttt{\{david.beauchemin, richard.khoury\}@ift.ulaval.ca},\\
\textsuperscript{\textdaggerdbl}\texttt{zachary.gagnon.1@ulaval.ca}
}

\begin{document}
\maketitle
\begin{abstract}
Large Language Models (LLMs) perform outstandingly in various downstream tasks, and the use of the Retrieval-Augmented Generation (RAG) architecture has been shown to improve performance for legal question answering \cite{nuruzzaman2020intellibot, louis2024interpretable}.
However, there are limited applications in insurance questions-answering, a specific type of legal document.
This paper introduces two corpora: the Quebec Automobile Insurance Expertise Reference Corpus and a set of 82 Expert Answers to Layperson Automobile Insurance Questions.
Our study leverages both corpora to automatically and manually assess a GPT4-o, a state-of-the-art LLM, to answer Quebec automobile insurance questions.
Our results demonstrate that, on average, using our expertise reference corpus generates better responses on both automatic and manual evaluation metrics.
However, they also highlight that 
LLM QA is unreliable enough for mass utilization in critical areas. 
Indeed, our results show that between 5\% to 13\% of answered questions include a false statement that could lead to customer misunderstanding.
\end{abstract}

\section{Introduction}
\label{sec:intro}
To protect their financial situation and property, vehicle owners and homeowners need to buy property damage insurance.
However,  most people have little to no proper knowledge of insurance products, and rely on insurance representatives to help them properly select and comprehend these products \citep{loiamf, planamf}.
As a result, in order to protect the public, insurance regulators, such as the \guillemet{Autorité des marchés financiers} (AMF) in Quebec, make sure that insurance representatives are well-trained and educated, and that insurers properly inform their customers \citep{amfmission}. 

However, customers are increasingly interested in buying insurance products online \citep{cefrio}. This change impacts how an insurer can adequately inform their customer. Traditionally, customers buy products in person or through phone insurance representatives, which
allows an insurance expert to help the customer understand the different products and buy the correct one. 
With an online sale, customers are left to gather information by themselves \citep{memoireamfloi141, RCCAQ}.
Moreover, insurance is regulated locally, which means that insurance products, coverages and laws are different from one jurisdiction to the next. 
Consequently, while many resources are available online, such as \guillemet{infoassurance} \citep{infoassmission}, only the limited set of resources from one's own locality are applicable, and customers must take care not to get information from elsewhere.

The rapid progress in natural language processing and the growing availability of insurance data present unprecedented opportunities to bridge the gap between people and insurance knowledge. 
For instance, legal text summarization \citep{shukla2022legal} holds the potential to simplify complex legal documents for laypeople. Similarly, insurance question-answering (QA) could offer affordable, expert-like assistance to non-expert customers.

To this end, we present an end-to-end approach aimed at generating high-quality answers to Quebec automobile insurance questions. Our methodology harnesses the popular \guillemet{Retrieval-Augmented Generation} (RAG) approach.
The main contributions of this work are therefore:

\begin{enumerate}[leftmargin=*, noitemsep]
    \item The creation and release of a Quebec Automobile Insurance Expertise References Corpus\footnote{\href{https://github.com/GRAAL-Research/quebec-insurance-rag-corpora}{https://github.com/GRAAL-Research/quebec-insurance-rag-corpora}}; 
    \item The creation and release of a corpus of 82 Expert Answers to Laypeople Automobile Insurance Questions\footnote{\href{https://github.com/GRAAL-Research/quebec-insurance-rag-corpora}{https://github.com/GRAAL-Research/quebec-insurance-rag-corpora}}; 
    \item A set of experiments to assess the performance of GPT-4o, a state-of-the-art LLM, on our QA corpus, including a manual evaluation of the generated answers.
\end{enumerate}

This paper is outlined as follows: first, we study the relevant questions-answering legal RAG research and its related corpora in \autoref{sec:related}. 
Then, we propose our corpora in \autoref{sec:corpora}, and in \autoref{sec:metho}, \autoref{sec:exp} and \autoref{sec:res} we present a set of experiments
aimed at evaluating the performances of GPT-4o at answering Quebec automobile insurance questions. 
Finally, in \autoref{sec:conclusion}, we conclude and discuss our future work.

\section{Related Work}
\label{sec:related}

\paragraph{Legal-Domain QA RAG}
The advent of large language models (LLMs) has led to advances in many previously arduous tasks, such as in the application of the RAG concept in QA tasks, which has attracted a great deal of research interest in recent years \citep{pipitone2024legalbenchragbenchmarkretrievalaugmentedgeneration}. Answering legal questions has always been more complex due to the inherent difficulties of exploiting specialized texts that stem from handing specialized terminology \citep{wiratunga2011case} and intricate sentence structures \cite{katz2023natural}. 
Recently, \citet{louis2024interpretable} has presented an end-to-end methodology to generate answers to any statutory law question leveraging a RAG architecture, along with a long-form legal question answering dataset comprising 1,868 expert-annotated legal questions in French. 
Likewise, the insurance sector, with its complex documents and nuanced information, could benefit from these advancements.    
Consequently, although research is mainly focused on the legal field, there is also a growing interest in the insurance sector, including for insurance RAG. 
\citet{nuruzzaman2020intellibot} have presented a chatbot that generates accurate and contextual responses by identifying intentions and entities while ensuring semantic relevance and meaning of responses. It is trained on domain-specific datasets to understand insurance-specific terms and information. It notably uses RAG strategies to generate responses. 
Likewise, \citet{na2022insurance} focuses on a single-turn dialogue covering insurance QA on a Korean dataset to respond to insurance customers. 

\paragraph{Legal and Insurance Corpora}
The number of datasets available in the legal and insurance domains has increased in recent years \cite{martinez2023survey, cui2023survey}. 
One example is CUAD \cite{hendrycks1cuad}, a dataset for legal contract review that includes 13,000 human annotations.
The first insurance QA dataset was proposed by 
\citet{feng2015applying}, and consists in 16,889 question-answer pairs; they also conducted experiments to assess different approaches at answering insurance questions.
More recently, \citet{butler-2023-open-australian-legal-dataset} have proposed a corpus of 2,124 synthetic question-answer pairs concerning Australian law.
The corpus was generated using GPT-4 and the Open Australian Legal Corpus, but the answers were not reviewed by an insurance expert.
However, as of yet, no such corpus exists for automobile insurance questions.

\section{Corpora}
\label{sec:corpora}
This section describes the two corpora we created for our work: our French corpus of automobile insurance expertise references documentation for the Province of Quebec (Canada), and our French corpus of 82 layperson questions about Quebec automotive insurance and their expert answers and annotations.
First, we will describe our process for creating each corpus\footnote{We also discuss the risk of data leakage in our Limitations section.} and then present some key statistics.

\subsection{Corpus Creation}
\subsubsection{Quebec Automobile Insurance Expertise Reference Corpus}
This corpus is composed of a set of documents extracted from seven official and reliable online sources about automobile insurance in Quebec.
These sources have been selected in partnership with a Canadian insurance company.
They have been divided into the following four categories: 

\begin{itemize}[leftmargin=*, noitemsep]
    \item The \textbf{Laws} category includes two pieces of provincial legislation related to Quebec automobile insurance. 
    The first one is the \textit{Loi sur l'assurance automobile} \citep{loiassauto}, which establishes the regulations governing insurers and insureds in Quebec. 
    The second one is the \textit{Code de sécurité routière}
    \citep{loicoderoute}, and it governs the use of all motorized vehicles and pedestrians on public roads to ensure safety.
    \item The \textbf{F.P.Q. 1} category includes the manually extracted Quebec mandatory-approved automobile insurance contracts \citep{amf_form, Beauchemin2023RISC}. The F.P.Q. 1 is divided into civil liability and property damage. Optional coverages are described in endorsements. We have included one realistic synthetic contract that includes all available endorsements.
    \item The \textbf{Insurance Regulator Educative Resources} category includes informative resources from the AMF, Quebec's regulatory body for financial and insurance products and services \citep{amfmission}. 
    We included its educational information related to automobile insurance for customers.
    \item The \textbf{Domain-Specific Educative Resources} category includes educative resources from four insurance domain organizations.
    They all propose various educational resources to the public through their online blog.
    The first, the \textit{Chambre de l'Assurance de Dommages}, is the regulatory body that oversees the training and ethics of insurance agents, brokers and claims adjusters \citep{chadmission}.
    The second, the \textit{Groupement des Assureurs Automobiles}, is the association of all home and car insurance insurers in Quebec. It oversees and develops various mechanisms to improve the property damage system \citep{gaamission}.
    The third, \textit{Éducaloi}, is a non-profit organization created by the Quebec Ministry of Justice that informs the public on legal matters, such as insurance products \citep{educaloi}.
    Lastly, \textit{Infoassurances} is an insurance information website created by the Insurance Bureau of Canada and the \textit{Groupement des Assureurs Automobiles} for the purpose of \guillemet{properly informing customers about property insurance} \citep{infoassmission}.
\end{itemize}

We have selected 21 online documents from these sources that focus on the subject of \guillemet{automobile insurance}.
The documents can be pieces of legislation, legal insurance documents, informative resources, or informative blog articles.
The content of each document has been manually extracted and cleaned to remove trailing whitespace, along with paragraphs that are either \guillemet{replaced} or \guillemet{repealed} in a piece of legislature.

\subsubsection{Corpus of Expert Answers to Laypeople's Automobile Insurance Questions}
This corpus comprises a set of French questions and answers related to automotive insurance in Quebec. They were manually extracted from highly-reliable sources that were selected in partnership with a Canadian insurance company, like for the previous references corpus. Our selected sources are divided into the following four categories: 

\begin{itemize}[leftmargin=*, noitemsep]
    \item The \textbf{Quebec Insurance Company FAQ} category includes question-answer pairs taken from the FAQ web pages of four insurers', namely, Beneva \citep{beneva}, Desjardins Assurances \citep{desjardins}, Belairdirect \citep{belair} and Sonnet \citep{sonnet}.
    These insurers have been selected based on two selection criteria.
    First, they must sell automotive insurance in Quebec.
    Second, the questions in their FAQ must not overlap with those of other selected insurers. 
    For example, Intact Assurance's \citep{intact} FAQ is identical to Belairdirect's, since both companies belong to the same corporation\footnote{\href{https://www.intactfc.com/en}{https://www.intactfc.com/en}}, and therefore that insurer was excluded. 
    \item The \textbf{Regulator} category includes insurance professional practice examination questions and answers from the regulator \citep{amffaq}.
    \item The \textbf{Domain-Specific Educative Resources} category includes question-answer pairs available through two educative resources and blogs from insurance sector organizations, namely the \textit{Chambre de l'Assurance de Dommages} and \textit{Infoassurances}.
    These two sources are also used as reference sources. We have carefully ensured no overlap between the extracted questions and the extracted reference content from these sources.
    \item The \textbf{Quebec Public Automobile Insurance Plan} category includes question-answer pairs from the Quebec government agency responsible for the automobile insurance plan that covers all bodily injuries \citep{saaq}.
\end{itemize}

We extracted 82 question-answer pairs from these sources, along with a category for each pair.
Seven categories were extracted from the sources; each question is related to one of the following categories:

\begin{itemize}[leftmargin=*, noitemsep]
    \item \textbf{Legal Obligations} are questions related to the insuree's and insurer's legal obligation. For example, it could be a question about the minimum amount of civil liability insurance required.
    \item \textbf{Civil Liability Coverage} are questions related to civil liability coverage. This could be for example a question about how a civil liability claim works.
    \item \textbf{Property Coverage} are questions related to at-fault accidents and the scope of property damage protection. For example, there could be a question about how to file an at-fault claim.
    \item \textbf{Endorsement} are questions related to any endorsements in insurance contracts. For example, it could be a question about the protections found in an endorsement.
    \item \textbf{Terms and Conditions} are questions related to an insurance contract's general terms and conditions. It could be a question about the consequence of a customer not paying their premium for instance.
    \item \textbf{General} are questions related to the general elements of the insurance sector. One example could be a question about why insurance companies use credit scores during the insurance proposal.
    \item \textbf{Public Automobile Insurance Plan} are questions related to bodily injury coverage offered by the public automobile insurance plan in Quebec. This for example could be a question about the program coverage and exclusions.
\end{itemize}

\subsection{Corpora Analysis}
\autoref{tab:statistics} presents some key statistics of our French corpora and similar English insurance QA corpora introduced in \autoref{sec:related}. 
For the English insurance QA corpora, we have used their latest official version\footnote{\href{https://github.com/shuzi/insuranceQA}{https://github.com/shuzi/insuranceQA}, \href{https://huggingface.co/datasets/umarbutler/open-australian-legal-qa}{https://huggingface.co/datasets/umarbutler/open-australian-legal-qa}}.
All statistics were computed using SpaCy's latest language-specific tokenizer \cite{Honnibal_spaCy_Industrial-strength_Natural_2020}. 
They exclude new lines (\verb|\n|), whitespaces, punctuations and some special characters (\verb|<|, \verb|>|, \texttt{|} and \verb|$|). 
Moreover, to evaluate the reading complexity level of the contracts, we compute readability scores using the frequently used Flesch-Kincaid formula \cite{flesch1948readability}. It computes a score using a scale from 0 (hardest) to 100 (easier) to assess the readability level.
We will first analyze our reference corpus and then compare our QA corpus with similar corpora using \autoref{tab:statistics}.

\subsubsection{Our References Corpus Analysis}
In \autoref{tab:statistics} (left side), we see that all four sources share relatively similar statistics.
Indeed, the average number of lexical words (LW), average sentence lengths (both), and average number of sentences are relatively similar.
Moreover, since legal documents are known to be complex and lengthy and to use specialized vocabulary \cite{katz2023natural}, we can see that the average number of tokens, lexical richness and average Flesch-Kincaid score are lower than the two other types of documents.

\subsubsection{Question-Answering Corpora Comparison}
We can see in \autoref{tab:statistics} (right side) that our QA corpus shares similar patterns to the other corpora.
Indeed, for all corpora, the questions use less than half the vocabulary size as the answers and are half as long in terms of tokens, LW, number of sentences, and average sentence length as the answers.
They are also easier to read than the answers based on the Flesch-Kincaid score.
However, ours is significantly smaller compared to other similar corpora due to its nature.
Indeed, the other two similar corpora focus on the broader insurance domain.
For example, Insurance QA includes questions about all types of insurance (property, life, and health) throughout the USA. 
In contrast, our corpus focuses on a single insurance product for a single province in Canada.

\begin{table*}
    \tiny
    \resizebox{\textwidth}{!}{%
    \begin{tabular}{@{}lccccc|cccccc@{}}
    \toprule
       & \multicolumn{5}{c}{References Corpus}                & \multicolumn{2}{|c}{Our QA Corpus} & \multicolumn{2}{c}{Australian Legal QA} & \multicolumn{2}{c}{Insurance QA} \\
       & Laws  & F.P.Q. 1 & Regulator & Sector & Avg   & Questions              & Answers             & Questions           & Answers           & Questions       & Answers       \\ \midrule
        Number of QA pair                                         & N/A   & N/A      & N/A       & N/A    & N/A   & \multicolumn{2}{c}{82}                       & \multicolumn{2}{c}{2,124}               & \multicolumn{2}{c}{16,889}      \\
        Vocabulary size                                                        & 4,638 & 1,751    & 1,038     & 1,029  & 6,201 & 367                    & 950                 & 6,657               & 13,583            & 3,658           & 19,355        \\
        Avg number of tokens                                                   & 89.41 & 87.43    & 115.52    & 109.83 & 90.60 & 14.45                  & 57.98               & 26.03               & 85.99             & 7.36            & 99.98         \\
        Avg number of LW                                                       & 38.95 & 42.14    & 51.2      & 49.71  & 40.20 & 6.68                   & 25.67               & 14.57               & 44.6              & 4.03            & 45.91         \\
        Avg number of sentence                                                 & 4.13  & 8.12     & 7.33      & 7.12   & 4.96  & 1.24                   & 3.00                & 1.41                & 3.12              & 1.00            & 5.42          \\
        Avg sentence length (tokens) & 21.37 & 11.81    & 17.05     & 16.67  & 19.56 & 12.22                  & 20.86               & 21.09               & 31.22             & 7.32            & 19.57         \\
        Avg sentence length (LW)     & 9.25  & 5.89     & 7.61      & 7.51   & 8.61  & 5.59                   & 9.34                & 11.74               & 16.52             & 4.02            & 9.12          \\
        Lexical richness                                                       & 0.11  & 0.18     & 0.37      & 0.36   & 0.10  & 0.48                   & 0.37                & 0.21                & 0.14              & 0.05            & 0.02          \\
        Avg Flesch-Kincaid score                                               & 46.67 & 56.45    & 61.41     & 65.6   & 49.31 & 73.66                  & 60.19               & 55.8                & 46.1              & 71.25           & 66.78         \\ \bottomrule
        \end{tabular}%
    }
    \caption{Aggregate statistics of our French corpora and similar English insurance QA corpora introduced in \autoref{sec:related}. \guillemet{Avg} stands for average, \guillemet{LW} for lexical words.}
    \label{tab:statistics}
\end{table*}

\section{QA Methodology}
\label{sec:metho}
\begin{figure*}[ht!]
    \centering
    \includegraphics[width=0.95\linewidth]{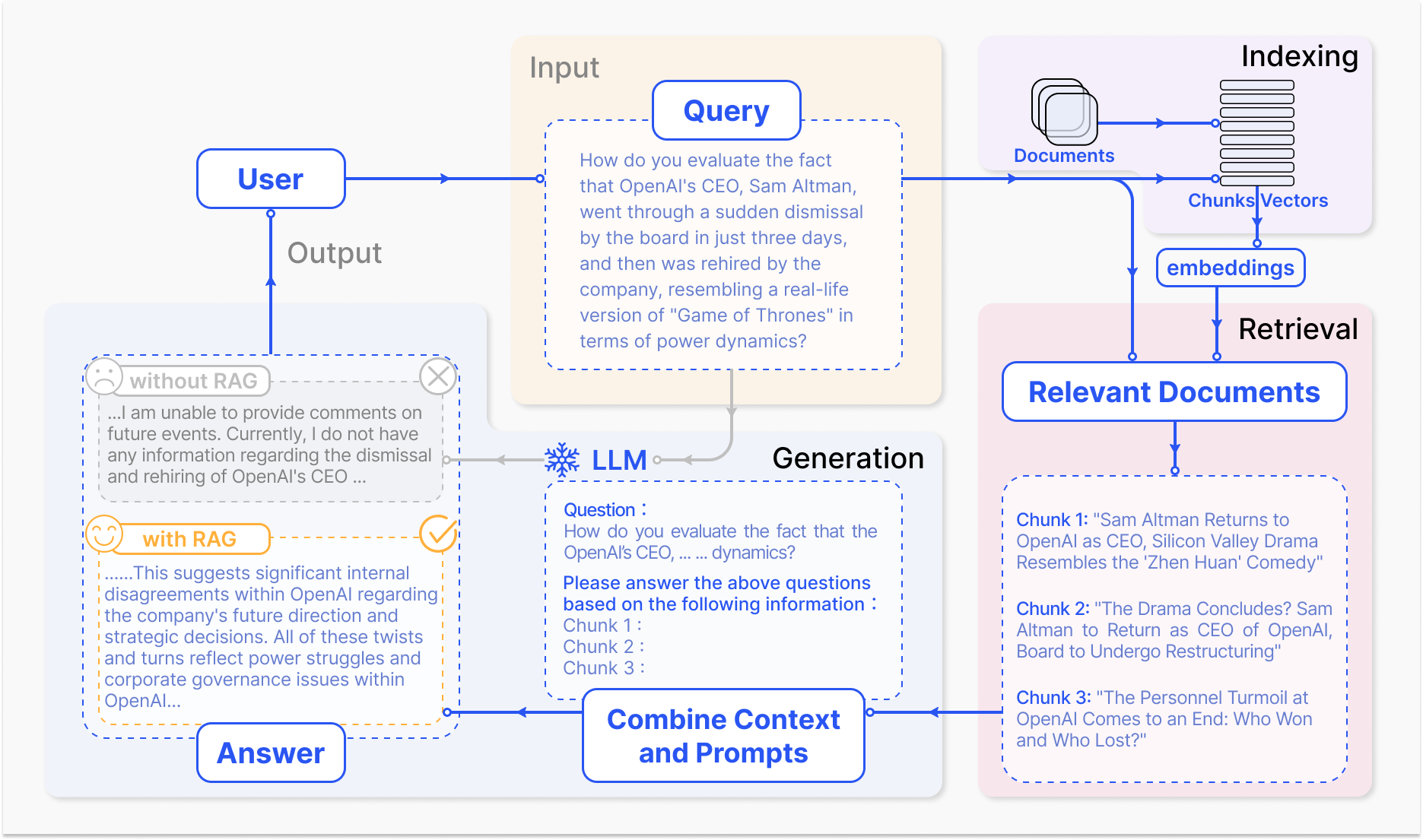}
    \caption{A representative instance of the 3-steps RAG process applied to question answering. 1) \hlc{purplerag}{Indexing}: Documents are split into chunks, and encoded into vectors in a vector database. 2) \hlc{redrag}{Retrieval}: Retrieve the Top k chunks most relevant to the question based on semantic similarity. 3) \hlc{bleurag}{Generation}: Input the original question and the retrieved chunks together into LLM to generate the final answer. The illustration is taken from \citet{gao2023retrieval}.}
    \label{fig:rag}
\end{figure*}

This section details our methodology for leveraging a large language model (LLM) to answer insurance questions.
Our choice of architecture is similar to \citet{louis2024interpretable}, \citet{ajmi2024revolutionizing}, and \citet{wiratunga2024cbr}. We use a RAG architecture to inject domain expertise into an LLM generation for QA.
Like the previous authors, our RAG architecture is inspired by the concept of \guillemet{advanced RAG} \cite{gao2023retrieval}, an architecture that adds a pre- and post-retrieval steps to the traditional processing. Our architecture was built using \texttt{LangChain} \cite{Chase_LangChain_2022}, a Python framework that consolidates the various components of the RAG architecture.
As illustrated in \autoref{fig:rag}, first, a retriever selects a small subset of insurance documents from our reference corpus (\hlc{redrag}{red}), some relevant to the question and some not. 
Then, a generator conditions its answer on the subset of articles returned by the retriever (\hlc{bleurag}{blue}). We describe these two components in details in the following subsections.

\subsection{Retriever}
The function of our retriever component is to extract from our reference corpus portions of texts, such as sentences or paragraphs, that are relevant to a question and to present them at the forefront of the returned results. It is a two-step operation consisting of pre-processing and retrieval steps.

\subsubsection{Pre-Processing}
During the pre-processing step, all our documents in the reference corpus go through a two-step pre-processing stage to prepare our document for our retrieval algorithm. 
The first step is to split the document into smaller chunks of text (i.e. chunking).
Based on the best practices for RAG in \citet{wang2024searching}, we use a fixed chunk size of 500 characters which gives optimal performance for document search since it standardizes their size for better similarity search results.
Moreover, legal documents are similar to the financial reports of \citet{yepes2024financial} because they use a standard structure to present their content. 
For example, laws are divided in chapters composed of articles relevant to their subject, which are in turn composed of sub-articles related to the main article. 
We thus process the documents using a parent-child split function to capture this structure.
However, the complete chunk is also supplied for generation when the similarity function is performed during retrieval on the child-split.
Namely, if a sub-article is extracted as a relevant text, the main article's text chunk will be provided, not only the sub-article.

The second step is to encode all chunks into dense embedding representation for retrieval.
To do so, we use \texttt{text-embedding-ada-002} \cite{Greene_Neelakantan_Weng_Sanders_2022}, a 1,536 dimension multilingual all-purpose embeddings model proposed by OpenAI.
This embedding model has proven successful in the insurance field \cite{mohanan2024competitive}.

\subsubsection{Retrieval}
\label{ref:retrieval}
The retrieval step seeks to retrieve a subset of articles using an algorithm that leverages dense word embeddings (i.e. \texttt{text-embedding-ada-002}) for retrieval. 
Our retrieval process is a 3-step process that uses the question as a query. 
First, the question is encoded using the retriever embedding model.
Second, using our dense retriever, we retrieve the \texttt{top-5} relevant documents from the reference corpus using cosine similarity to measure the semantic similarity between the query and each document. 
Third, we merge all relevant reference documents using a context compressor \cite{cheng2024xrag}. 
This compressor calls an LLM for each reference using the extracted document (context), the user query, and a formatted prompt that specifies the compressor's task. 
With this prompt, the LLM is asked to return only the relevant part of the context given the query and, if needed, reformulate the context in certain difficult-to-understand cases, as is sometimes the case with technical legal texts. 
The compressor reduces the context size, thus keeping the prompt size within an acceptable range, in order to prevent certain issues. 
Indeed, it is known that excessively large prompts can degrade the quality of answer generation \cite{levy2024tasktokensimpactinput}. 
Moreover, a lost-in-the-middle effect can cause a language model to omit information contained in the middle portion of the prompt \cite{liu2024lost}.
This approach helps merge content from different sources to create a better-contextualized reference document for an LLM to generate an answer \cite{wang2024searching}.

\subsection{Generation}
Our generator's goal is to formulate an exhaustive and concise answer to an automotive insurance question based on the information provided by the retrieval process.
Our generator uses \texttt{GPT-4o}, the latest OpenAI LLM model.
The prompt is constructed using the question and the context obtained from the retrieved reference documents, along with specific task instructions designed to guide the LLM in formulating a comprehensive and accurate answer.

As shown in \autoref{fig:prompts} and \autoref{fig:prompts2}, we have used two prompts for our experiment. The first (\autoref{fig:prompts}) is a zero-shot prompt where the LLM is simply asked to answer the question. The second (\autoref{fig:prompts2} is a domain-specific prompt that gives additional information to support the LLM. 
In each prompt, \texttt{\{input\}} corresponds to the question, and \texttt{\{context\}} to the retrieved references.

\begin{figure}[ht!]
        \tiny
        \begin{subfigure}[b]{\linewidth}
        \centering
            \begin{tikzpicture}[scale=1, every node/.style={transform shape}]
            
            \node[rectangle, rounded corners, draw=tgb1, fill=tgb1!80, text width=0.9\linewidth, align=left, inner sep=1.2ex] (prompt) {Répondez à la question suivante EN FRANÇAIS.};
            
            \node[rectangle, rounded corners, draw=tgb4, fill=tgb4, below=0.1cm of prompt, text width=0.9\linewidth, align=left, inner sep=1.2ex] (input) {Voici la question: \{input\}.};
            \end{tikzpicture}
            \caption{Basic zero-shot prompt adapted from \citet{kew2023bless} followed by the input question to respond to.}
            \label{fig:promptFR}
        \end{subfigure}
        
        \begin{subfigure}[b]{\linewidth}
        \centering
            \begin{tikzpicture}[scale=1, every node/.style={transform shape}]
    
            \node[rectangle, rounded corners, draw=tgb1, fill=tgb1!80, text width=0.9\linewidth, align=left, inner sep=1.2ex] (prompt) {Answer the following question IN FRENCH.};
            
            \node[rectangle, rounded corners, draw=tgb4, fill=tgb4, below=0.1cm of prompt, text width=0.9\linewidth, align=left, inner sep=1.2ex] (input) {Here's the question: \{input\}.};
            \end{tikzpicture}
            \caption{Translation of the prompt presented in \autoref{fig:promptFR}.}
            \label{fig:promptEN}
        \end{subfigure}
    \caption{Zero-shot prompt used for text generation. 
    \hlc{tgb1!80}{Blue} boxes contain the task instructions. \hlc{tgb4}{Yellow} boxes contain the prefix for the model to continue.}
    \label{fig:prompts}
\end{figure}

\begin{figure}[ht!]
        \tiny
        \begin{subfigure}[b]{\linewidth}
        \centering
            \begin{tikzpicture}[scale=1, every node/.style={transform shape}]
            
            \node[rectangle, rounded corners, draw=tgb1, fill=tgb1!80, text width=0.9\linewidth, align=left, inner sep=1.2ex] (prompt) {Vous êtes un expert en assurances automobile dans le domaine de l’assurance de dommages. Vous répondez à des questions EN FRANÇAIS liés à l’assurance automobile AU QUÉBEC. Vous utilisez le contexte fourni ci-bas. Répondez EN PHRASES COMPLÊTES et soyez concis.};
            
            \node[rectangle, rounded corners, draw=tgb4, fill=tgb4, below=0.1cm of prompt, text width=0.9\linewidth, align=left, inner sep=1.2ex] (input) {Voici la question: \{input\};\\Voici le contexte : \{contexte\}.};
            \end{tikzpicture}
            \caption{Domain-specific prompt with prompt engineering (i.e. role, task, domain of application) adapted from \citet{kew2023bless} followed by the input question to respond to.}
            \label{fig:prompt2FR}
        \end{subfigure}
        
        \begin{subfigure}[b]{\linewidth}
        \centering
            \begin{tikzpicture}[scale=1, every node/.style={transform shape}]
    
            \node[rectangle, rounded corners, draw=tgb1, fill=tgb1!80, text width=0.9\linewidth, align=left, inner sep=1.2ex] (prompt) {You are an automobile insurance expert in the property and casualty insurance field. You are answering questions in FRENCH related to automobile insurance in QUEBEC. Use the context provided below. Answer in FULL PHRASES and be concise.};
            
            \node[rectangle, rounded corners, draw=tgb4, fill=tgb4, below=0.1cm of prompt, text width=0.9\linewidth, align=left, inner sep=1.2ex] (input) {Here's the question: \{input\};\\Here's the context: \{context\}.};
            \end{tikzpicture}
            \caption{Translation of the prompt presented in \autoref{fig:prompt2FR}.}
            \label{fig:prompt2EN}
        \end{subfigure}
    \caption{Prompt used for text generation. 
    \hlc{tgb1!80}{Blue} boxes contain the task instructions. \hlc{tgb4}{Yellow} boxes contain the prefix for the model to continue.}
    \label{fig:prompts2}
\end{figure}

\section{Experiments}
\label{sec:exp}
The goal of our experiments is to assess whether an LLM can adequately answer technical questions with complex answers, namely Quebec insurance questions, with or without a RAG architecture. To achieve this, we conduct experiments to automatically and manually evaluate six approaches.

\subsection{Experimentation Setup}
\paragraph{Baseline} For our baseline, we use our zero-shot prompt to assess GPT-4o out-of-the-box capabilities to answer Quebec insurance questions. We label it \texttt{Zero-shot} in our result tables.

\paragraph{RAG Architecture Approaches} For our other five experiments, we use our RAG architecture described in \autoref{sec:metho} and the domain-specific prompt, with an increasing number of reference sources. 
Namely, we start with an approach that uses no references. The difference between this approach and the baseline is only prompt engineering.
Then, we incrementally add in reference sources. 
The next approach only uses the \textbf{Laws} source, the following adds the \textbf{F.P.Q. 1}, then we add the AMF reference, and finally we add the educative resources to use all four references. We label these five approaches, \texttt{No references}, \texttt{Laws}, \texttt{Laws, F.P.Q. 1}, \texttt{Laws, F.P.Q. 1, AMF} and \texttt{All references} respectively.

\subsection{Evaluation}
\subsubsection{Automatic Evaluation}
Following \citet{chen2019evaluating}, we evaluate the accuracy of machine-generated answers compare to reference answers using three $N$-grams based metrics: BLEU-\{1, 4, AVG\} \cite{papineni2002bleu}, ROUGE-\{1, 2, L\} (F1-Score) \cite{lin2004looking} and METEOR \cite{banerjee2005meteor} scores.
We also use two deep similarity metrics that measure the similarity between a machine-generated text and a reference document to compute \guillemet{how semantically related are those two documents} using words embedding: BERTScore \cite{zhangbertscore}, and MeaningBERT \cite{Beauchemin_MeaningBERT_assessing_meaning_2023}. Each metric uses a slightly different approach to compute this similarity. The first feeds the machine- and human-generated documents separately through a BERT model, then computes a token-by-token alignment between the documents using pairwise cosine similarity. The second, MeaningBERT, uses a fine-tuned pre-trained BERT model train to predict how similar two documents are; the model aims to maximize its correlation with human evaluation.
We report the results averaged over five restarts from different random seeds.

\subsubsection{Manual Evaluation}
To discern the strengths and shortcomings of our generator with or without using an RAG architecture, we conduct a detailed manual analysis of all question-answer pairs.
Inspired by \citet{chartier2024evaluation} and \citet{baray2024modele}, we, in partnership with our insurance partner, have developed an evaluation guide with an exam-like setup to evaluate each pair.
Based on the expert answers, we defined a set of criteria, or key elements that a machine-generated answer must include.
To evaluate each criterion, we developed a grading scale inspired by the one used by \citet{chartier2024evaluation} and \citet{baray2024modele}; this scale is presented in \autoref{tab:grading}. In case of a false answer to a criterion, we penalize the score with a negative point since an erroneous answer could mislead the customer or hinder their understanding of an insurance product. On the other hand, a complete answer to a criterion results in the maximum score of 2 points. 
In total, 288 criteria have been extracted from the human answers.
On average, each question has 3.51 criteria with a standard deviation of 1.75. The maximum grade a system can receive is $288 \times 2 = 576$ points, and the lowest is $-288$ points when a system always gives a false answer.

Since we ran 5 runs of each setup with random restarts, we randomly select one of the five for our manual evaluation. 
One of the authors, with ten years of experience in Quebec Insurance, conducted the evaluation.
\autoref{ann:annotation} presents the evaluation interface used for our evaluation (in French).
During the evaluation, the evaluator is randomly presented with a randomly-selected generated answer from one of the six experimental setups, and he does not know which approach he evaluate.

\begin{table}
    \footnotesize
    \begin{tabular}{cp{0.38\textwidth}}
    \toprule
    Grade & Description     \\\midrule
    -1             & The system gives a false answer for the criterion i. For example, an answer states that civil liability covers property damage on the insured car if the owner is responsible, which is false. \\
    0              & The system does not give a proper answer to the criterion i or give an answer at all.                                                                                                                  \\
    1              & The system gives an incomplete answer to the criterion i.                                                                                                                     \\
    2              & The system gives a complete answer to the criterion i.                                   \\\bottomrule                                                                                    
    \end{tabular}%
    \caption{Our evaluation grading scale to evaluates a machine-generated answer using a set of criteria.}
    \label{tab:grading}
\end{table}

\section{Results}
\label{sec:res}
In this section, we present and discuss both our quantitative and qualitative results.
We also have conducted an ablation study that also use each source individually in \autoref{ann:ablation}.

\subsection{Quantitative Results}
The left-hand side of \autoref{table:res} presents the results of the automatic metrics averaged over the five random restarts, with \textbf{bolded} value indicating the best score.
First, we observe that, for all automatic metrics, on average, the \texttt{All references} approach outperforms other methods.
Moreover, this method's BLEU, ROUGE, and METEOR scores indicate that it gives answers using a vocabulary similar to that of humans in the ground truth.  
These scores are 40\% to 300\% higher than the zero-shot baseline approach.
It shows that using all our references greatly improves the LLM's ability to answer 
questions properly.
However, surprisingly, the second best approach is the \texttt{No reference} approach, which outperforms approaches using the same prompt along with a subset of our references. 
We hypothesize that using Laws and other juridical documents confuses the LLM and generates longer text that are penalized by automatic $N$-grams metrics.
We will explore and discuss this in the following section.

A second observation is that the approach with the highest variation in performance over the five restarts is \texttt{All references}.
Indeed, this approach's standard deviation is the highest of all setups, and is nearly three times higher than the lowest one.
It indicates that using this approach can also yield suboptimal generations.

Finally, to further assess our approaches' performance, we report the two best approaches z-test significance test in \autoref{table:z_test}. Our null hypothesis is that the pair of approaches have equal performances, meaning that values smaller or greater than $|1.96|$ allow us to reject the null hypothesis with $\alpha = 0.05$. A positive value means that the \texttt{No references} model (left) performs significantly better than the \texttt{All references} (right), and a negative value means the opposite.
We can see that for most metrics, \texttt{All references} has a significantly better performance compared to \texttt{No references}; we can conclude that \texttt{All references} is better than \texttt{No references}.

\begin{table*}
    \footnotesize   
    \centering
    \resizebox{\textwidth}{!}{%
        \begin{tabular}{p{0.175\textwidth}ccccccccc|cc}
        \toprule
            \multicolumn{1}{c}{\multirow{2}{*}{}} & \multicolumn{3}{c}{ROUGE}                                             & \multicolumn{1}{c}{\multirow{2}{*}{BERTScore}} & \multicolumn{1}{c}{\multirow{2}{*}{MeaningBERT}} & \multicolumn{3}{c}{BLEU}                                                & \multicolumn{1}{c}{\multirow{2}{*}{METEOR}} & \multirow{2}{*}{\begin{tabular}[c]{@{}c@{}} Exam \\Score (\%)\end{tabular}} & \multirow{2}{*}{\begin{tabular}[c]{@{}c@{}} False\\Statement\end{tabular}}\\
            \multicolumn{1}{c}{}                  & \multicolumn{1}{c}{1} & \multicolumn{1}{c}{2} & \multicolumn{1}{c}{L} & \multicolumn{1}{c}{}                            & \multicolumn{1}{c}{}                                                       & \multicolumn{1}{c}{Average} & \multicolumn{1}{c}{1} & \multicolumn{1}{c}{4} & \multicolumn{1}{c}{}   &            &    \\\midrule

        \texttt{Zero-shot} & 0.27$\pm$0.10 & 0.09$\pm$0.06 & 0.16$\pm$0.06 & 66.93$\pm$4.33 & 71.42$\pm$11.21 & 4.10$\pm$4.25 & 21.28$\pm$10.63 & 1.39$\pm$2.63 & 24.23$\pm$7.61 & 27.43&34\\
        \texttt{No references} & 0.35$\pm$0.09 & 0.14$\pm$0.08 & 0.22$\pm$0.07 & 71.40$\pm$4.43 & 78.17$\pm$10.62 & 7.06$\pm$6.16 & 33.63$\pm$14.63 & 3.05$\pm$5.00 & 27.02$\pm$9.68 & 32.29&20\\
        \texttt{Laws} & 0.32$\pm$0.1 & 0.12$\pm$0.08 & 0.20$\pm$0.07 & 70.743$\pm$4.54 & 77.05$\pm$11.11 & 6.177$\pm$5.76 & 31.276$\pm$15.05 & 2.76$\pm$5.49 & 26.29$\pm$9.68& 27.78&20\\
        \texttt{Laws, F.P.Q. 1} & 0.32$\pm$0.11 & 0.13$\pm$0.11 & 0.21$\pm$0.1 & 70.29$\pm$5.1 & 75.44$\pm$11.0 & 6.73$\pm$7.47 & 30.40$\pm$15.02 & 3.30$\pm$6.26 & 27.05$\pm$10.42& 29.51&19\\
        \texttt{Laws, F.P.Q. 1, AMF} & 0.33$\pm$0.11 & 0.14$\pm$0.11 & 0.21$\pm$0.1 & 70.89$\pm$5.59 & 76.91$\pm$10.63 & 7.62$\pm$8.02 & 31.76$\pm$16.0 & 3.93$\pm$7.12 & 27.76$\pm$10.78& 34.20&18\\
        \texttt{All references} & \textbf{0.375$\pm$0.14} & \textbf{0.18$\pm$0.15} & \textbf{0.25$\pm$0.14} & \textbf{71.99$\pm$5.9} & \textbf{78.87$\pm$10.17} & \textbf{10.68$\pm$11.71} & \textbf{33.77$\pm$16.66} & \textbf{5.98$\pm$9.91} & \textbf{33.61$\pm$14.69}& \textbf{51.74}&14\\
        \bottomrule
        \end{tabular}%
    }
    \caption{Automatic metrics (left) average and one standard deviation over the five restarts on our questions-answering corpus and manual (right) evaluation using our evaluation guide. The best score is \textbf{bolded}.}
    \label{table:res}
\end{table*}

\begin{table*}
    \footnotesize
    \centering
    \resizebox{\textwidth}{!}{%
        \begin{tabular}{p{0.20\textwidth}ccccccccc|c}
        \toprule
            \multicolumn{1}{c}{\multirow{2}{*}{}} & \multicolumn{3}{c}{ROUGE}                                             & \multicolumn{1}{c}{\multirow{2}{*}{BERTScore}} & \multicolumn{1}{c}{\multirow{2}{*}{MeaningBERT}} & \multicolumn{3}{c}{BLEU}                                                & \multicolumn{1}{c}{\multirow{2}{*}{METEOR}} & \multirow{2}{*}{\begin{tabular}[c]{@{}c@{}} Exam\\Score (\%)\end{tabular}}\\
            \multicolumn{1}{c}{}                  & \multicolumn{1}{c}{1} & \multicolumn{1}{c}{2} & \multicolumn{1}{c}{L} & \multicolumn{1}{c}{}                                                      & \multicolumn{1}{c}{}                            & \multicolumn{1}{c}{AVG} & \multicolumn{1}{c}{1} & \multicolumn{1}{c}{4} & \multicolumn{1}{c}{}   &                \\\midrule

        \texttt{No references}/\texttt{All references}  & \textbf{-3.25} & \textbf{-3.94} & \textbf{-3.28} & -1.57 & \textbf{-2.47} & \textbf{-3.60} & \textbf{-4.00} & \textbf{-3.49} & \textbf{-2.96} & \textbf{-2.52}\\\bottomrule
        \end{tabular}%
    }
    \caption{Z-test significance test of our two bests approaches (\textbf{bold} value are rejected null hypothesis with $\alpha = 0.05$).}
    \label{table:z_test}
\end{table*}

\subsection{Qualitative Results}
The right-hand side of \autoref{table:res} also presents the manual grading obtained using our evaluation guide, with \textbf{bolded} value indicating the best score.
Once again, we observe that \texttt{All references} approach outperforms other methods, achieving a score  nearly double that of the baseline method.

Moreover, \texttt{No references} scores are higher than approaches that use a portion of the references corpus.
This seems to indicate that responses from partial references are not just longer but are also incomplete. 
Indeed, we observed that using legal documents generates longer responses, but the generated answers tend to be of lower quality. 
For example, to the question \guillemet{What is the recommended amount of civil liability insurance I should carry when driving outside Quebec?} (translated), the \texttt{Laws} model answers with the definition of civil liability 
instead of responding with the recommended amount of 2 million dollars.
In contrast, the \texttt{No references} approach answers with the correct amount. 
It is likely due to data leakage: GPT-4o might have been trained using some of our references and memorized the correct answer. By forcing a different context from incomplete references, the LLM seems to forget or overwite that information. 

An interesting situation occurred with the question \guillemet{I was injured in a car accident. What should I do?} (translated). All evaluated generations take the questions literally and assume the driver has just been injured, and thus propose steps to secure the insuree such as \guillemet{call an ambulance}.
In contrast, the ground truth specified the administrative steps to proceed with a bodily injury claim. 
It shows that, in this case, LLM cannot infer the actual context of the question.

Another interesting situation is whether or not the model abstains from answering in cases where the context is unknown or the information to respond to the question is unavailable for the model.
In none of the cases we examined, the model abstained from answering the question. It always strived to be as helpful as possible. However, while sufficient, our prompt could be enhanced to further boost performance. By adapting it to generate better responses and prevent the model from responding when uncertain, we hypothesize that we could improve its performance by improving the prompt.

Moreover, in many cases, without specific references to Quebec insurance specifications, the response contained French insurance information. For example, the \texttt{No References} model responded to many questions with specific details of automobile insurance with France-based examples such as civil liability coverage. This pattern disappeared with the addition of the references.

Finally, we can see that the zero-shot approach generates the lowest grade and the highest number of false answers. 
This highlights the risk of using an out-of-the-box LLM to generate technical answers with precise answer elements, as in our situation. 
It also highlights that using our RAG approach with our references corpus can lower this risk substantially.
While the risk of false answers remain present, it is a better way for
consumers get easier access to insurance expertise.

No analysis was done as to how the system performs when the question is out of the context of references -- does it hallucinate an answer, does it abstain from answering? Would be important to classify what kinds of questions can be answered by the system in order to put guard-rails on it.

\subsection{Discussions}
Evaluation of RAG systems typically relies on automatic generation N-Grams metrics \cite{yu2024evaluation}. 
As our results highlight, these metrics provide interesting insight into model performance. Such insight was used to steer the development of the solution. However, the legal field and documents are known to be lengthy and complex \cite{beauchemin2020generating, Beauchemin2023RISC}. Thus, we are skeptical that only relying on this type of metrics is sufficient to develop robust systems; these metrics display an incomplete illustration of the system's response quality and cannot properly capture the legal and misinformation risks they pose to the public.
Indeed, ROUGE and BLEU have been criticized for lacking semantic capabilities or correlating weakly with human judgment \cite{reiter2018structured, tay2019red, Beauchemin_MeaningBERT_assessing_meaning_2023}.
Moreover, more recent approaches that leverage Transformer-based architecture, such as BertScore, have yet to be shown to achieve a strong correlation with human judgment \cite{Beauchemin_MeaningBERT_assessing_meaning_2023}.
For this reason, many RAG applications now focus on human evaluation \cite{yu2024evaluation}. 
However, such an evaluation procedure is labour-intensive and costly, especially in specialized fields such as the legal domain.
Our primary results show that one can use automatic metrics during development to steer one project. However, human evaluation should evaluate the final system qualitatively to assess a system's performance and risk properly, particularly in sensitive fields such as the legal domain.

\section{Conclusion and Future Works}
\label{sec:conclusion}
In conclusion, this paper introduced two new corpora: a Quebec automobile insurance expertise reference corpus and a corpus of expert answers to laypeople automobile insurance questions.
To generate answers to the questions in our second corpus, we leverage an RAG architecture that uses our reference corpus.
We experimented with six approaches: a zero-shot that did not use the RAG architecture, an RAG architecture without references, and four models that incrementally use more of our reference corpus. 
Our results demonstrate that, on average, using our complete reference corpus generates better responses based on both automatic and manual evaluation metrics.
Our results show that between 5\% to 13\% of generated answers include a false statement that could mislead a customer, indicating that LLM-based technical and sensitive QA is not yet robust enough for mass utilization by the public.

In our future works, we plan to extend the references corpus to include AMF proprietary documents, such as their insurance representative training manual, and increase the number of expert-answered questions. 
Moreover, we would also like to experiment with other LLMs, and to conduct a real-life evaluation using real insurance customers.
Finally, we plan to improve performance with prompt engineering and LLM fine-tuning.

\section*{Limitations}

First, despite our efforts to make our systems more factually grounded using Quebec insurance references, our proposed framework remains at risk of generating hallucinations in its answers, as shown in \autoref{table:res}. 

Second, since our reference documents are available online, it is possible that GPT-4o and other LLMs could have been trained with 
some or all of our reference documents. Thus, the results we obtained may include some overfitting, which could make it difficult to generalize to unseen data.

Third, our study is limited to monolingual French documents and QA, and to a single application domain. Though we expect our results to be consistent in other languages and domains, we did not study that question. 

Fourth, we acknowledge that our prompt might be considered too simplistic; our focus was not to rabbit-hole ourselves with prompt engineering but instead study the quantitative and qualitative capabilities of out-of-the-box solutions and minimalist RAG to assess the limitations of such technology.

Finally, consistent with prior studies \citep{krishna2021hurdle, xu2023critical, louis2024interpretable}, we observe that conventional automatic metrics may not accurately mirror answer quality, leading to potential misinterpretations.

\section*{Ethical considerations}
As highlighted by \citet{beauchemin2020generating}, the premature deployment of legal NLP solutions, such as an insurance RAG system for the Quebec Insurance domain, poses a tangible risk to laypeople, who may uncritically rely on the answers it provides and thus inadvertently exacerbate their circumstances. Indeed, a layperson might use this kind of innovation as a viable source of information. Thus, the quality of the response needs to be as precise as possible.
Furthermore, the use of AI in the legal field poses significant risks because of the presence of bias in corpora and the systems where many might be considered illegal \cite{bender2021dangers, beauchemin:hal-03736828}.
We are committed to limiting the use of our dataset strictly to research purposes to ensure the responsible development of legal aid technologies and limit the risk of illegal, biased use.

\section*{Hardware \& Librairies} Computations are performed on two 12 GO NVIDIA GTX 1080 TI and with proprietary OpenAI LLM and embeddings model using their API; one experimentation over the six approaches cost around 30 USD, the overall OpenAI budget was 1,050 USD.

\section*{Acknowledgements}
This research was made possible thanks to the support of a Canadian insurance company, NSERC research grant RDCPJ 537198-18 and FRQNT doctoral research grant. 
We thank the reviewers for their comments regarding our work.

\appendix
\section{Evaluation Annotation Interface}
\label{ann:annotation}

\begin{figure*}
    \centering
    \includegraphics[width=1\linewidth]{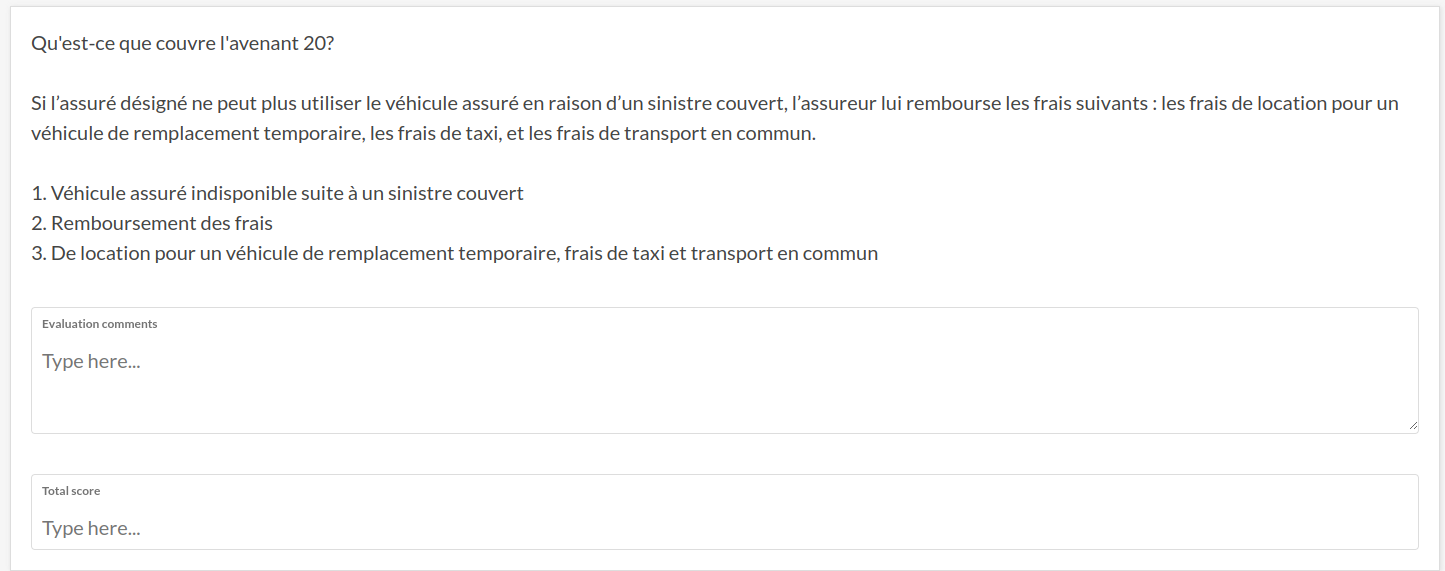}
    \caption{The Prodigy annotation interface (in French) used for evaluation.}
    \label{fig:clf_annotation}
\end{figure*}

\autoref{fig:clf_annotation} presents the evaluation interface used for our evaluation (in French). It is a custom adaptation of the Prodigy annotation tool~\cite{prodigy_montani_honnibal}. 

\section{Ablation Study}
\label{ann:ablation}

\autoref{tab:ablation} presents the ablation study based on the references used for the RAG, namely using only one source reference instead of the cumulative approach. Our results show that using the cumulative approach yields better results than using only one.
We did not conduct the manual evaluation of our ablation study.

\begin{table*}
    \footnotesize   
    \centering
    \resizebox{\textwidth}{!}{%
        \begin{tabular}{p{0.175\textwidth}ccccccccc}
        \toprule
            \multicolumn{1}{c}{\multirow{2}{*}{}} & \multicolumn{3}{c}{ROUGE}                                             & \multicolumn{1}{c}{\multirow{2}{*}{BERTScore}} & \multicolumn{1}{c}{\multirow{2}{*}{MeaningBERT}} & \multicolumn{3}{c}{BLEU}                                                & \multicolumn{1}{c}{\multirow{2}{*}{METEOR}}\\
            \multicolumn{1}{c}{}                  & \multicolumn{1}{c}{1} & \multicolumn{1}{c}{2} & \multicolumn{1}{c}{L} & \multicolumn{1}{c}{}                            & \multicolumn{1}{c}{}                                                       & \multicolumn{1}{c}{Average} & \multicolumn{1}{c}{1} & \multicolumn{1}{c}{4} \\\midrule

        \texttt{All references} & 0.375$\pm$0.14 & 0.18$\pm$0.15 & 0.25$\pm$0.14 & 71.99$\pm$5.9 & 78.87$\pm$10.17 & 10.68$\pm$11.71 & 33.77$\pm$16.66 & 5.98$\pm$9.91 & 33.61$\pm$14.69\\
        \midrule
        \texttt{F.P.Q. 1} & 0.212$\pm$0.10 & 0.09$\pm$0.06 & 0.16$\pm$0.06 & 66.93$\pm$4.33 & 71.42$\pm$11.21 & 4.10$\pm$4.25 & 21.28$\pm$10.63 & 1.39$\pm$2.63 & 24.23$\pm$7.61 \\
        \texttt{AMF} & 0.210$\pm$0.19 & 0.13$\pm$0.08 & 0.19$\pm$0.06 & 70.46$\pm$4.54 & 76.17$\pm$10.45 & 7.06$\pm$6.16 & 33.63$\pm$14.63 & 3.05$\pm$5.00 & 27.02$\pm$9.68\\
        \texttt{Educative Resources} & 0.240$\pm$0.08 & 0.13$\pm$0.07 & 0.19$\pm$0.09 & 71.26$\pm$4.34 & 77.55$\pm$10.22 & 7.36$\pm$5.16 & 33.17$\pm$12.62 & 3.44$\pm$5.01 & 27.38$\pm$9.68\\
        \bottomrule
        \end{tabular}%
    }
    \caption{Automatic metrics average and one standard deviation over the five restarts on our questions-answering corpus of our ablation study.}
    \label{tab:ablation}
\end{table*}

\bibliography{anthology,custom}

\begin{thebibliography}{64}
\expandafter\ifx\csname natexlab\endcsname\relax\def\natexlab#1{#1}\fi

\bibitem[{Ajmi(2024)}]{ajmi2024revolutionizing}
Ayyoub Ajmi. 2024.
\newblock \href {https://irlaw.umkc.edu/faculty_works/949} {{Revolutionizing
  Access to Justice: The Role of AI-Powered Chatbots and Retrieval-Augmented
  Generation in Legal Self-Help}}.
\newblock In \emph{The Brief}, volume 53-10.

\bibitem[{AMF()}]{amf_form}
Autorité des marchés~financiers AMF.
\newblock {AMF approved forms}.
\newblock Accessed online (14-08-2024)
  \url{https://lautorite.qc.ca/en/professionals/insurers/automobile-insurance/amf-approved-forms}.

\bibitem[{AMF(2018)}]{memoireamfloi141}
Autorité des marchés~financiers AMF. 2018.
\newblock \href
  {https://lautorite.qc.ca/fileadmin/lautorite/grand_public/publications/professionnels/assemblee-nationale/20180118-memoire-pl141.pdf}
  {\emph{Mémoire présenté à la Commission des finances publiques sur le
  Projet de loi 141 : Loi visant principalement à améliorer l’encadrement
  du secteur financier, la protection des dépôts d’argent et le régime de
  fonctionnement des institutions financières}}.
\newblock Autorité des marchés financiers.

\bibitem[{AMF(2019)}]{planamf}
Autorité des marchés~financiers AMF. 2019.
\newblock \href
  {https://lautorite.qc.ca/fileadmin/lautorite/education-financiere/SQEF2019-plan-action_an.pdf}
  {\emph{{Québec Financial Education Strategy for 2019-2022 - Orientations and
  Action Plan}}}.
\newblock Autorité des marchés financiers.

\bibitem[{AMF(2024{\natexlab{a}})}]{amfmission}
Autorité des Marchés~Financiers AMF. 2024{\natexlab{a}}.
\newblock {Mission}.
\newblock Accessed online (14-08-2024)
  \url{https://lautorite.qc.ca/en/general-public/about-the-amf/mission}.

\bibitem[{AMF(2024{\natexlab{b}})}]{amffaq}
Autorité des Marchés~Financiers AMF. 2024{\natexlab{b}}.
\newblock {Practice Examination Questions}.
\newblock Accessed online (14-08-2024)
  \url{https://lautorite.qc.ca/en/becoming-a-professional/damage-insurance/examinations/practice-examination-questions}.

\bibitem[{Banerjee and Lavie(2005)}]{banerjee2005meteor}
Satanjeev Banerjee and Alon Lavie. 2005.
\newblock {METEOR: An Automatic Metric for MT Evaluation With Improved
  Correlation With Human Judgments}.
\newblock In \emph{Proceedings of the acl workshop on intrinsic and extrinsic
  evaluation measures for machine translation and/or summarization}, pages
  65--72.

\bibitem[{Baray et~al.(2024)Baray, Decrop, and Cliquet}]{baray2024modele}
J{\'e}r{\^o}me Baray, Alain Decrop, and G{\'e}rard Cliquet. 2024.
\newblock Mod{\`e}le standardis{\'e} d'{\'e}valuation des ia
  g{\'e}n{\'e}ratives en soutien {\`a} la recherche marketing: Test de chatgpt.
\newblock In \emph{Colloque international de l'Association Tunisienne de
  Marketing}.

\bibitem[{Beauchemin et~al.(2020)Beauchemin, Garneau, Gaumond, D{\'e}ziel,
  Khoury, and Lamontagne}]{beauchemin2020generating}
David Beauchemin, Nicolas Garneau, Eve Gaumond, Pierre-Luc D{\'e}ziel, Richard
  Khoury, and Luc Lamontagne. 2020.
\newblock {Generating Intelligible Plumitifs Descriptions: Use Case Application
  with Ethical Considerations}.
\newblock In \emph{Proceedings of the International Conference on Natural
  Language Generation}, pages 15--21.

\bibitem[{Beauchemin and Khoury(2023)}]{Beauchemin2023RISC}
David Beauchemin and Richard Khoury. 2023.
\newblock {RISC: Generating Realistic Synthetic Bilingual Insurance Contract}.
\newblock \emph{Proceedings of the Canadian Conference on Artificial
  Intelligence}.
\newblock Https://caiac.pubpub.org/pub/k18zu6c9.

\bibitem[{Beauchemin and Monty(2022)}]{beauchemin:hal-03736828}
David Beauchemin and Marie-Claire Monty. 2022.
\newblock \href {https://hal.science/hal-03736828} {{La discrimination en
  intelligence artificielle est-elle suffisamment encadr{\'e}e ?}}
\newblock Preprint.

\bibitem[{Beauchemin et~al.(2023)Beauchemin, Saggion, and
  Khoury}]{Beauchemin_MeaningBERT_assessing_meaning_2023}
David Beauchemin, Horacio Saggion, and Richard Khoury. 2023.
\newblock \href {https://doi.org/10.3389/frai.2023.1223924} {{MeaningBERT:
  Assessing Meaning Preservation Between Sentences}}.
\newblock \emph{Frontiers in Artificial Intelligence}, 6.

\bibitem[{Belairdirect(2024)}]{belair}
Belairdirect. 2024.
\newblock {FAQ}.
\newblock Accessed online (14-08-2024)
  \url{https://www.belairdirect.com/en/faq.html}.

\bibitem[{Bender et~al.(2021)Bender, Gebru, McMillan-Major, and
  Shmitchell}]{bender2021dangers}
Emily~M Bender, Timnit Gebru, Angelina McMillan-Major, and Shmargaret
  Shmitchell. 2021.
\newblock {On the Dangers of Stochastic Parrots: Can Language Models Be Too
  Big?}
\newblock In \emph{Proceedings of the ACM conference on fairness,
  accountability, and transparency}, pages 610--623.

\bibitem[{Beneva(2024)}]{beneva}
Beneva. 2024.
\newblock {FAQ - Car Insurance}.
\newblock Accessed online (14-08-2024)
  \url{https://www.beneva.ca/en/car-insurance/help}.

\bibitem[{Butler(2023)}]{butler-2023-open-australian-legal-dataset}
Umar Butler. 2023.
\newblock \href {https://doi.org/10.57967/hf/1479} {{Open Australian Legal
  QA}}.

\bibitem[{ChAD(2024)}]{chadmission}
Chambre de l'Assurance de~Dommages ChAD. 2024.
\newblock {About Us}.
\newblock Accessed online (14-08-2024) \url{https://chad.ca/en/about-us/}.

\bibitem[{Chartier et~al.(2024)Chartier, Dakkoune, Bourgeois, and
  Jean}]{chartier2024evaluation}
Mathieu~Alexandre Chartier, Nabil Dakkoune, Guillaume Bourgeois, and
  St{\'e}phane Jean. 2024.
\newblock {\'E}valuation des capacit{\'e}s de r{\'e}ponse de larges mod{\`e}les
  de langage (llm) pour des questions d'historiens.
\newblock In \emph{24{\`e}me conf{\'e}rence francophone sur l'Extraction et la
  Gestion des Connaissances}, 40, pages 155--166.

\bibitem[{Chase(2022)}]{Chase_LangChain_2022}
Harrison Chase. 2022.
\newblock \href {https://github.com/langchain-ai/langchain} {{LangChain}}.

\bibitem[{Chen et~al.(2019)Chen, Stanovsky, Singh, and
  Gardner}]{chen2019evaluating}
Anthony Chen, Gabriel Stanovsky, Sameer Singh, and Matt Gardner. 2019.
\newblock {Evaluating Question Answering Evaluation}.
\newblock In \emph{Proceedings of the workshop on machine reading for question
  answering}, pages 119--124.

\bibitem[{Cheng et~al.(2024)Cheng, Wang, Zhang, Ge, Chen, Wei, Zhang, and
  Zhao}]{cheng2024xrag}
Xin Cheng, Xun Wang, Xingxing Zhang, Tao Ge, Si-Qing Chen, Furu Wei, Huishuai
  Zhang, and Dongyan Zhao. 2024.
\newblock {xRAG: Extreme Context Compression for Retrieval-augmented Generation
  with One Token}.
\newblock \emph{arXiv:2405.13792}.

\bibitem[{Claire et~al.(2018)Claire, Marie-Eve, René, Sébastien, Lova,
  Guillaume, Annie, and Marie-Guy}]{cefrio}
Bourget Claire, Lacombe Marie-Eve, Godbout René, Lanctôt Sébastien,
  Rajaobelina Lova, Ducharme Guillaume, Lavoie Annie, and Maynard Marie-Guy.
  2018.
\newblock \emph{{Assurance de dommages à l’ère du numérique}}.
\newblock Centre facilitant la recherche et l’innovation dans les
  organisations.

\bibitem[{Cui et~al.(2023)Cui, Shen, and Wen}]{cui2023survey}
Junyun Cui, Xiaoyu Shen, and Shaochun Wen. 2023.
\newblock {A Survey on Legal Judgment Prediction: Datasets, Metrics, Models and
  Challenges}.
\newblock \emph{IEEE Access}.

\bibitem[{{Desjardins Assurances}(2024)}]{desjardins}
{Desjardins Assurances}. 2024.
\newblock {Insurance FAQ}.
\newblock Accessed online (14-08-2024)
  \url{https://www.desjardins.com/qc/en/insurance/faq.html}.

\bibitem[{Feng et~al.(2015)Feng, Xiang, Glass, Wang, and
  Zhou}]{feng2015applying}
Minwei Feng, Bing Xiang, Michael~R Glass, Lidan Wang, and Bowen Zhou. 2015.
\newblock {Applying Deep Learning to Answer Selection: A Study and an Open
  Task}.
\newblock In \emph{IEEE workshop on automatic speech recognition and
  understanding}, pages 813--820. IEEE.

\bibitem[{Flesch(1948)}]{flesch1948readability}
Rudolf Flesch. 1948.
\newblock {A Readability Formula in Practice}.
\newblock \emph{Elementary English}, 25(6).

\bibitem[{GAA(2024)}]{gaamission}
Groupement des Assureurs~Automobiles GAA. 2024.
\newblock {About Us}.
\newblock Accessed online (14-08-2024) \url{https://gaa.qc.ca/en/who-are-we/}.

\bibitem[{Gao et~al.(2023)Gao, Xiong, Gao, Jia, Pan, Bi, Dai, Sun, and
  Wang}]{gao2023retrieval}
Yunfan Gao, Yun Xiong, Xinyu Gao, Kangxiang Jia, Jinliu Pan, Yuxi Bi, Yi~Dai,
  Jiawei Sun, and Haofen Wang. 2023.
\newblock {Retrieval-Augmented Generation for Large Language Models: A Survey}.
\newblock \emph{arXiv:2312.10997}.

\bibitem[{Greene et~al.(2022)Greene, Neelakantan, Weng, and
  Sanders}]{Greene_Neelakantan_Weng_Sanders_2022}
Ryan Greene, Arvind Neelakantan, Lilian Weng, and Ted Sanders. 2022.
\newblock \href {https://openai.com/index/new-and-improved-embedding-model}
  {{New and Improved Embedding Model}}.

\bibitem[{Hendrycks et~al.()Hendrycks, Burns, Chen, and Ball}]{hendrycks1cuad}
Dan Hendrycks, Collin Burns, Anya Chen, and Spencer Ball.
\newblock {CUAD: An Expert-Annotated NLP Dataset for Legal Contract Review}.
\newblock In \emph{Conference on Neural Information Processing Systems Datasets
  and Benchmarks Track}.

\bibitem[{Honnibal et~al.(2020)Honnibal, Montani, Van~Landeghem, and
  Boyd}]{Honnibal_spaCy_Industrial-strength_Natural_2020}
Matthew Honnibal, Ines Montani, Sofie Van~Landeghem, and Adriane Boyd. 2020.
\newblock \href {https://doi.org/10.5281/zenodo.1212303} {{SpaCy:
  Industrial-strength Natural Language Processing in Python}}.

\bibitem[{IBC and GAA(2024)}]{infoassmission}
Insurance Bureau of~Canada IBC and Groupement des Assureurs~Automobiles GAA.
  2024.
\newblock {Infoassurance - About Us}.
\newblock Accessed online (14-08-2024)
  \url{https://infoassurance.ca/en/utility-menu/about-us/}.

\bibitem[{{Intact Insurance}(2024)}]{intact}
{Intact Insurance}. 2024.
\newblock {FAQ}.
\newblock Accessed online (14-08-2024) \url{https://www.intact.ca/en/faq}.

\bibitem[{Johnson(2018)}]{RCCAQ}
Christopher Johnson. 2018.
\newblock {Projet de loi 141 et vente par internet: où en est le RCCAQ?}
\newblock
  \url{https://www.rccaq.com/cgi/page.cgi/_article_fr.html/Categories/Dans_la_mire/Projet_de_loi_141_et_vente_par_internet_o_en_est_le_RCCAQ_}.

\bibitem[{Katz et~al.(2023)Katz, Hartung, Gerlach, Jana, and
  Bommarito}]{katz2023natural}
Daniel~Martin Katz, Dirk Hartung, Lauritz Gerlach, Abhik Jana, and
  Michael~James Bommarito. 2023.
\newblock {Natural Language Processing in the Legal Domain}.
\newblock \emph{Available at SSRN 4336224}.

\bibitem[{Kew et~al.(2023)Kew, Chi, V{\'a}squez-Rodr{\'\i}guez, Agrawal,
  Aumiller, Alva-Manchego, and Shardlow}]{kew2023bless}
Tannon Kew, Alison Chi, Laura V{\'a}squez-Rodr{\'\i}guez, Sweta Agrawal, Dennis
  Aumiller, Fernando Alva-Manchego, and Matthew Shardlow. 2023.
\newblock {BLESS: Benchmarking Large Language Models on Sentence
  Simplification}.
\newblock \emph{arXiv:2310.15773}.

\bibitem[{Krishna et~al.(2021)Krishna, Roy, and Iyyer}]{krishna2021hurdle}
Kalpesh Krishna, Aurko Roy, and Mohit Iyyer. 2021.
\newblock \href {https://doi.org/10.18653/v1/2021.naacl-main.393} {{Hurdles to
  Progress in Long-form Question Answering}}.
\newblock In \emph{Proceedings of the Conference of the North American Chapter
  of the Association for Computational Linguistics: Human Language
  Technologies}, pages 4940--4957. Association for Computational Linguistics.

\bibitem[{Levy et~al.(2024)Levy, Jacoby, and
  Goldberg}]{levy2024tasktokensimpactinput}
Mosh Levy, Alon Jacoby, and Yoav Goldberg. 2024.
\newblock \href {http://arxiv.org/abs/arXiv:2402.14848} {{Same Task, More
  Tokens: the Impact of Input Length on the Reasoning Performance of Large
  Language Models}}.

\bibitem[{Lin and Och(2004)}]{lin2004looking}
Chin-Yew Lin and FJ~Och. 2004.
\newblock {Looking for a Few Good Metrics: ROUGE and Its Evaluation}.
\newblock In \emph{Ntcir workshop}.

\bibitem[{Liu et~al.(2024)Liu, Lin, Hewitt, Paranjape, Bevilacqua, Petroni, and
  Liang}]{liu2024lost}
Nelson~F Liu, Kevin Lin, John Hewitt, Ashwin Paranjape, Michele Bevilacqua,
  Fabio Petroni, and Percy Liang. 2024.
\newblock {Lost in the Middle: How Language Models Use Long Contexts}.
\newblock \emph{Transactions of the Association for Computational Linguistics},
  12:157--173.

\bibitem[{Louis et~al.(2024)Louis, van Dijck, and
  Spanakis}]{louis2024interpretable}
Antoine Louis, Gijs van Dijck, and Gerasimos Spanakis. 2024.
\newblock {Interpretable Long-Form Legal Question Answering With
  Retrieval-Augmented Large Language Models}.
\newblock In \emph{Proceedings of the AAAI Conference on Artificial
  Intelligence}, volume~38, pages 22266--22275.

\bibitem[{Martinez-Gil(2023)}]{martinez2023survey}
Jorge Martinez-Gil. 2023.
\newblock {A Survey on Legal Question--Answering Systems}.
\newblock \emph{Computer Science Review}, 48:100552.

\bibitem[{Mohanan(2024)}]{mohanan2024competitive}
Monisha Mohanan. 2024.
\newblock \emph{{Competitive Analysis of Embedding Models in
  Retrieval-Augmented Generation for Indian Motor Vehicle Law Chat Bots}}.
\newblock Ph.D. thesis, Dublin Business School.

\bibitem[{Montani and Honnibal(2018)}]{prodigy_montani_honnibal}
Ines Montani and Matthew Honnibal. 2018.
\newblock \href {https://prodi.gy/} {{Prodigy: A Modern and Scriptable
  Annotation Tool for Creating Training Data for Machine Learning Models}}.

\bibitem[{Na et~al.(2022)Na, Kim, and Cho}]{na2022insurance}
Seon-Ok Na, Young-Min Kim, and Seung-Hwan Cho. 2022.
\newblock {Insurance Question Answering via Single-turn Dialogue Modeling}.
\newblock In \emph{Proceedings of the Second Workshop on When Creative AI Meets
  Conversational AI}, pages 35--41.

\bibitem[{Nuruzzaman and Hussain(2020)}]{nuruzzaman2020intellibot}
Mohammad Nuruzzaman and Omar~Khadeer Hussain. 2020.
\newblock {IntelliBot: A Dialogue-Based Chatbot for the Insurance Industry}.
\newblock \emph{Knowledge-Based Systems}, 196:105810.

\bibitem[{Papineni et~al.(2002)Papineni, Roukos, Ward, and
  Zhu}]{papineni2002bleu}
Kishore Papineni, Salim Roukos, Todd Ward, and Wei-Jing Zhu. 2002.
\newblock {BLEU: A Method for Automatic Evaluation of Machine Translation}.
\newblock In \emph{Proceedings of the annual meeting of the Association for
  Computational Linguistics}, pages 311--318.

\bibitem[{Pipitone and
  Alami(2024)}]{pipitone2024legalbenchragbenchmarkretrievalaugmentedgeneration}
Nicholas Pipitone and Ghita~Houir Alami. 2024.
\newblock \href {http://arxiv.org/abs/arXiv: 2408.10343} {{LegalBench-RAG: A
  Benchmark for Retrieval-Augmented Generation in the Legal Domain}}.

\bibitem[{{Quebec}(2016)}]{loicoderoute}
{Quebec}. 2016.
\newblock \href
  {https://www.legisquebec.gouv.qc.ca/fr/document/lc/C-24.2/20160908} {Code de
  la sécurité routière 2016}.

\bibitem[{{Quebec}(2024)}]{loiassauto}
{Quebec}. 2024.
\newblock \href {https://www.legisquebec.gouv.qc.ca/fr/document/lc/A-25} {Loi
  sur l’assurance automobile 2024}.

\bibitem[{Reiter(2018)}]{reiter2018structured}
Ehud Reiter. 2018.
\newblock {A Structured Review of the Validity of BLEU}.
\newblock \emph{Computational Linguistics}, 44(3):393--401.

\bibitem[{RLRQ(2004)}]{loiamf}
Recueil des lois et des règlements du~Québec RLRQ. 2004.
\newblock {Act Respecting the Regulation of the Financial Sector}.

\bibitem[{SAAQ(2024)}]{saaq}
Société de l'Assurance Automobile du~Québec SAAQ. 2024.
\newblock {Québec’s Public Automobile Insurance Plan in Brief}.
\newblock Accessed online (14-08-2024)
  \url{https://saaq.gouv.qc.ca/en/traffic-accident/public-automobile-insurance-plan/in-brief}.

\bibitem[{Shukla et~al.(2022)Shukla, Bhattacharya, Poddar, Mukherjee, Ghosh,
  Goyal, and Ghosh}]{shukla2022legal}
Abhay Shukla, Paheli Bhattacharya, Soham Poddar, Rajdeep Mukherjee, Kripabandhu
  Ghosh, Pawan Goyal, and Saptarshi Ghosh. 2022.
\newblock \href {https://aclanthology.org/2022.aacl-main.77} {{Legal Case
  Document Summarization: Extractive and Abstractive Methods and their
  Evaluation}}.
\newblock In \emph{Proceedings of the Conference of the Asia-Pacific Chapter of
  the Association for Computational Linguistics and the International Joint
  Conference on Natural Language Processing}, pages 1048--1064. Association for
  Computational Linguistics.

\bibitem[{Sonnet(2024)}]{sonnet}
Sonnet. 2024.
\newblock {Frequently Asked Questions}.
\newblock Accessed online (14-08-2024) \url{https://www.sonnet.ca/faqs}.

\bibitem[{Tay et~al.(2019)Tay, Joshi, Zhang, Karimi, and Wan}]{tay2019red}
Wenyi Tay, Aditya Joshi, Xiuzhen~Jenny Zhang, Sarvnaz Karimi, and Stephen Wan.
  2019.
\newblock {Red-Faced ROUGE: Examining the Suitability of ROUGE for Opinion
  Summary Evaluation}.
\newblock In \emph{Proceedings of the Annual Workshop of the Australasian
  Language Technology Association}, pages 52--60.

\bibitem[{Wang et~al.(2024)Wang, Wang, Gao, Zhang, Wu, Xu, Shi, Wang, Li, Qian
  et~al.}]{wang2024searching}
Xiaohua Wang, Zhenghua Wang, Xuan Gao, Feiran Zhang, Yixin Wu, Zhibo Xu,
  Tianyuan Shi, Zhengyuan Wang, Shizheng Li, Qi~Qian, et~al. 2024.
\newblock {Searching for Best Practices in Retrieval-Augmented Generation}.
\newblock \emph{arXiv:2407.01219}.

\bibitem[{Wiratunga et~al.(2024)Wiratunga, Abeyratne, Jayawardena, Martin,
  Massie, Nkisi-Orji, Weerasinghe, Liret, and Fleisch}]{wiratunga2024cbr}
Nirmalie Wiratunga, Ramitha Abeyratne, Lasal Jayawardena, Kyle Martin, Stewart
  Massie, Ikechukwu Nkisi-Orji, Ruvan Weerasinghe, Anne Liret, and Bruno
  Fleisch. 2024.
\newblock {CBR-RAG: Case-Based Reasoning for Retrieval Augmented Generation in
  LLMs for Legal Question Answering}.
\newblock In \emph{International Conference on Case-Based Reasoning}, pages
  445--460. Springer.

\bibitem[{Wiratunga and Ram(2011)}]{wiratunga2011case}
Nirmalie Wiratunga and Ashwin Ram. 2011.
\newblock \emph{{Case-Based Reasoning Research and Development}}.
\newblock Springer.

\bibitem[{Xu et~al.(2023)Xu, Song, Iyyer, and Choi}]{xu2023critical}
Fangyuan Xu, Yixiao Song, Mohit Iyyer, and Eunsol Choi. 2023.
\newblock \href {https://aclanthology.org/2023.acl-long.181} {{A Critical
  Evaluation of Evaluations for Long-form Question Answering}}.
\newblock In \emph{Proceedings of the Annual Meeting of the Association for
  Computational Linguistics}, pages 3225--3245. Association for Computational
  Linguistics.

\bibitem[{Yepes et~al.(2024)Yepes, You, Milczek, Laverde, and
  Li}]{yepes2024financial}
Antonio~Jimeno Yepes, Yao You, Jan Milczek, Sebastian Laverde, and Leah Li.
  2024.
\newblock {Financial Report Chunking for Effective Retrieval Augmented
  Generation}.
\newblock \emph{arXiv:2402.05131}.

\bibitem[{Yu et~al.(2024)Yu, Gan, Zhang, Tong, Liu, and Liu}]{yu2024evaluation}
Hao Yu, Aoran Gan, Kai Zhang, Shiwei Tong, Qi~Liu, and Zhaofeng Liu. 2024.
\newblock {Evaluation of Retrieval-Augmented Generation: A Survey}.
\newblock \emph{arXiv:2405.07437}.

\bibitem[{Zhang et~al.(2019)Zhang, Kishore, Wu, Weinberger, and
  Artzi}]{zhangbertscore}
Tianyi Zhang, Varsha Kishore, Felix Wu, Kilian~Q Weinberger, and Yoav Artzi.
  2019.
\newblock {BERTScore: Evaluating Text Generation with BERT}.
\newblock In \emph{International Conference on Learning Representations}.

\bibitem[{{Éducaloi}(2024)}]{educaloi}
{Éducaloi}. 2024.
\newblock {Governance}.
\newblock Accessed online (14-08-2024)
  \url{https://educaloi.qc.ca/en/governance/}.

\end{thebibliography}
\bibliographystyle{acl_natbib}

\end{document}